# Time Series Forecasting Using a Hybrid Deep Learning Method: A Bi-LSTM Embedding Denoising Auto Encoder Transformer

Sahar Koohfar, Wubeshet Woldemariam

**Time series data is a prevalent form of data found in various fields. It consists of a series of measurements taken over time. Forecasting is a crucial application of time series models, where future values are predicted based on historical data. Accurate forecasting is essential for making well-informed decisions across industries. When it comes to electric vehicles (EVs), precise predictions play a key role in planning infrastructure development, load balancing, and energy management. This study introduces a BI-LSTM embedding denoising autoencoder model (BDM) designed to address time series problems, focusing on short-term EV charging load prediction. The performance of the proposed model is evaluated by comparing it with benchmark models like Transformer, CNN, RNN, LSTM, and GRU. Based on the results of the study, the proposed model outperforms the benchmark models in four of the five-time steps, demonstrating its effectiveness for time series forecasting. This research makes a significant contribution to enhancing time series forecasting, thereby improving decision-making processes.**

## I. INTRODUCTION

Time series data is a widespread form of data found in various fields, such as engineering, medicine, finance, and economics. It involves a series of measurements taken over a period of time. The analysis of such data has contributed to the development of statistical and machine learning techniques. The growth of data and the emergence of new data sources has led to a greater need for machine learning models that are scalable and capable of handling complex time series problems. A crucial application of such models is forecasting, which involves predicting future values of a time series based on historical data. An accurate forecast is essential for making informed decisions across a variety of industries.

Several domains face the challenge of time series forecasting, including solar power forecasting [1], weather forecasting [2], traffic flow prediction [3], and energy systems, particularly regarding EV charging load forecasting. Considering the popularity of electric vehicles (EVs) as a more environmentally friendly alternative to traditional fossil fuel vehicles, individuals are increasingly switching from internal combustion engines (ICEs) to electric vehicles. Nevertheless, this transition has implications for the electricity grid, as the increased use of electric vehicles will adversely affect its stability. The random and volatile nature of EV charging loads can cause fluctuations in peak power, frequency, voltage, and overall energy demand, potentially straining the power grid. It is crucial to accurately predict EV charging loads to effectively manage this impact and ensure the grid remains stable and resilient. Prediction of EV charging load poses several challenges due to the nonlinear behavior relations, as well as the presence of both short-term and long-term patterns. The complex nature of EV charging load necessitates a deep learning approach, which can capture the intricate relationships between the input features and the target variable.

Accurate prediction of EV charging load is essential for evaluating the impact of EVs on the power grid and developing and operating a highly penetrable power system that promotes the adoption of EVs. Consequently, it is becoming increasingly necessary to develop reliable methods for predicting the charging loads of electric vehicles. As a result of accurate predictions, informed decisions can be made concerning infrastructure development, load balancing, and energy management, among other factors, to ensure a smooth integration of electric vehicles into the grid. In addition to addressing the challenges posed by EV charging loads, accurate prediction of EV charging load will contribute to the development of a sustainable and resilient power grid for future generations.

In this paper, we introduce a deep learning model designed to address time series problems, with a particular focus on solving the challenging problem of short-term EV charging load prediction. The main contributions of this study are:
• The Bi-LSTM denoising method is developed for time series forecasting. It provides a more robust and comprehensive understanding of the data, enabling more informed decision-making and analysis.
• The hybrid proposed method, BDT, uses a Bi-LSTM embedding algorithm. This algorithm allows for more accurate and targeted representation of structural elements in the data by aligning forward and backward paths simultaneously. This results in a more intricate and multidimensional portrayal of the data.
• In hybrid BDT, a denoising auto-encoder is utilized to eliminate noise and errors present in the input data, which improves the quality and reliability of predictions. Simultaneous use of cropping and denoising techniques improves prediction accuracy and reduces errors. The reduction of noise and errors in input data leads to more accurate and reliable predictions, even in unpredictable or challenging situations.

• The strengths and limitations of both Bi-LSTM and denoising auto encoder transformer architectures have been thoroughly analyzed, and we have identified an opportunity for merging these two models, thereby maximizing their unique advantages, and minimizing their respective drawbacks. Despite the promise of this approach, its application in the context of time series prediction of EV charging load demand has yet to be explored.

Our study aims to improve the accuracy of EV charging prediction models by integrating recent advancements in the Bi-LSTM and denoising auto encoder transformer architectures. We believe that this will advance the field of time series prediction techniques and lead to more informed decision-making when it comes to the development of EV charging infrastructure.

## II. LITERATURE REVIEW

This section examines the literature review of related research studies conducted by different researchers. Multiple approaches have been developed over the years to address the challenge of time series forecasting. Key techniques consist of statistical approaches, as well as deep learning methods.

Time series forecasting models can be classified into two main categories: statistical models and deep learning models. Statistical approaches focus on examining historical patterns and relationships within the data. On the other hand, deep learning approaches have the ability to automatically learn complex patterns from the data, which makes them particularly useful for handling non-linear and non-stationary time series [4].

With the aid of advanced techniques, researchers harnessed the inherent sequential characteristics of time series data. Among these methods were recurrent neural networks (RNNs) and their numerous variations, including long-short-term memory (LSTM), Bi-directional long-short memory (Bi-LSTM) and gated recurrent units (GRU). Moreover, convolutional neural networks (CNNs), originally developed for the analysis of images, have proven invaluable for analyzing time series data [5-11].

While RNNs, LSTMs, Bi-LSTMs, and CNNs have proven effective for time series forecasting, these models have limitations when dealing with datasets with long dependencies due to their sequential processing. To solve this problem, a transformer-based solution has been developed. Known for its exceptional performance in computer vision, speech recognition, and NLP applications, the transformer architecture has gained recognition as a highly effective sequence modeling technique [12-15]. Recently, there has been a notable increase in the use of transformer-based solutions for time series analysis, as they have shown superior performance compared to other deep learning models in time series forecasting. Both short-term and long-term time series prediction tasks can be performed by them with remarkable accuracy [16, 17].

A transformer's main strength lies in its multi-head self-attention mechanism, which enables it to capture semantic correlations between elements in long sequences of words. Transformers are able to extract and leverage essential patterns and dependencies within time series data using this mechanism. The application of transformer extensions, as well as hybrid approaches combining transformers with other techniques, has recently demonstrated superior performance in forecasting time series [18-20].

Although there have been significant advancements in time series forecasting, even the most accurate models are prone to making erroneous predictions. The conventional solution for improving forecasting accuracy is to provide more training data, but this is often a limited resource. Alternatively, given the success of denoising techniques in image generation, these techniques have shown success to enhance time series forecasting accuracy [14].

Recently, hybrid neural networks have become popular and proven to be effective in improving accuracy compared to traditional deep learning methods [21, 22]. By combining different deep learning techniques, these hybrid models excel in time series prediction. This success emphasizes the significance of exploring and combining the strengths of diverse deep learning approaches.

In this study, we have implemented a new deep learning model that incorporates the advantages of recent advancements in time series forecasting approaches. Our proposed model was compared to a benchmark deep learning model to evaluate its accuracy and effectiveness in forecasting time series data. The remaining sections of the paper are organized as follows: Section 3 provides a detailed explanation of the proposed model's structure. Section 4 covers the data and analysis part of the research, including the presentation of the model's performance results. Finally, Section 5 presents summary and conclusions of the study.

## 3. MATERICALS AND METHODS

### A. Bi-LSTM

Horchreiter and Schmidhuber introduced long short-term memory (LSTM) in 1997 as a solution to the problem of vanishing or expanding gradients [23]. LSTMs use gate mechanisms to determine which information should be retained or discarded during the learning process. They have three gated cell memories - input, forget, and output gates - which decide whether to store or delete data from the network model. The information with the highest significance is retained in the LSTM memory during training, while the rest is removed. Equations 1 through 5 provide the relationship between the gate mechanisms.

The where $i_t$, $f_t$, and $o_t$ are the input gate, forget gate, and output gate, respectively, at time $t$, $x_t$ is the input data at the time step $t$, and $h_{t-1}$ stands for the hidden layer in time $t$. $C_{t-1}$ and $C_t$ are cell states for the times $t-1$ and $t$, respectively, and $\tilde{C}_t$ is the internal memory unit. $W$ and $U$ denote the weight matrices associated with the corresponding gates.

The Bi-LSTM model is an extension of the LSTM models described above, consisting of two LSTM units that work in both directions to incorporate past and future information. The forward-moving LSTM receives past data from input sequences in the first round, while the backward-moving LSTM receives future data in the second round [24]. Figure 1 illustrates the structure of a Bi-LSTM model.

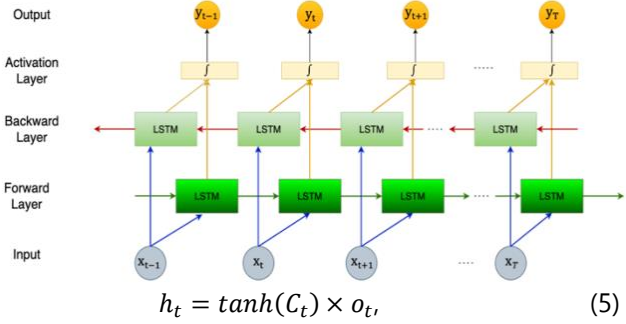

$$h_t = tanh(C_t) \times o_t, \tag{5}$$

**Fig. 1.** This is a sample of a figure caption.

In Bi-LSTM, the hidden layer contains both forward and backward tensors. The backward hidden state and forward hidden state at the time step t are shown as $\overrightarrow{h_t}$ and $\overleftarrow{h_t}$ respectively. For a hidden state at a specific time, step t is fed into the output layer, which is obtained by concatenating the forward and backward hidden states as follows:

$$h_t = [\overrightarrow{h_t}, \overleftarrow{h_t}] \tag{6}$$

The activation layer in Bi-LSTM network utilizes an activation function that is applied to the output of each Bi-LSTM layer. The activation function determines how the weighted sum of inputs is transformed into an output value, which then serves as the input for the subsequent layer in the network.

### B. Transformer

The first transformer architecture was published by Vaswani et al in 2017 [12]. This architecture uses self-attention to replace recurrent neural networks, which allows transformer models to selectively focus on important input data and improve the accuracy of time series forecasting [25]. Attention is particularly useful for extracting important information from large amounts of data, and transformers are able to access any point in the past without suffering from vanishing gradients, allowing them to detect long-term dependencies. The transformer model has an encoder-decoder structure, consisting of stacked encoder and decoder layers. The encoder layer contains a multi-head self-attention mechanism and a feed forward mechanism, while the decoder layer has three sub-layers: multi-head attention, masked multi-head attention, and feed forward mechanisms. The predictor input excludes subsequent positions with the use of multi-head attention. Each sublayer is surrounded by residual connections and layer normalization to speed up training and convergence, as shown in Figure 2.

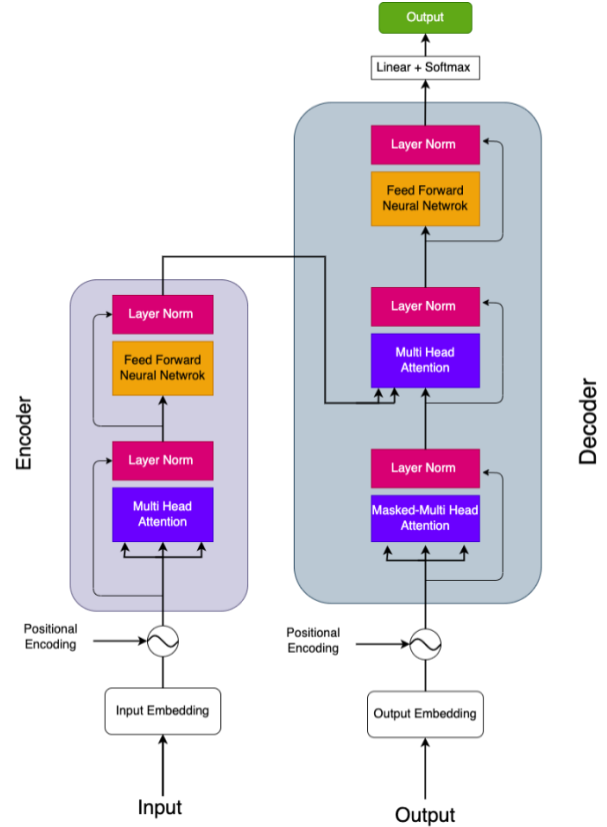

**Fig. 2.** Transformer architecture [17]

Sequences of time series data $(x_1, ..., x_n)$, are received by the encoder and converted into continuous representations $(z_1, ..., z_n)$. The decoder then uses these representations to calculate the output and generate symbols $(y_1, ..., y_m)$ one at a time. The transformer model is auto regressive, meaning it uses previous symbols as additional inputs. The output layer at each time step i, represented as $y_i$, is calculated as:

$$y_i = \sum_{j=1}^{n} a_{ij}(x_j W_V), \tag{7}$$

where $y_i$ is the updated $x_j$, and $a_{ij}$ is the attention score that measures the similarity between $x_i$ and $x_j$. $a_{ij}$ is calculated as:

$$a_{ij} = \frac{exp(e_{ij})}{\sum_{k=1}^{n} exp(e_{ik})}, \tag{8}$$

where $e_{ij}$ measures the combability of two linearly transformed input element $x_i$ and $x_j$:

$$e_{ij} = \frac{(x_i W_Q)(x_j W_K)^T}{\sqrt{h}}, \tag{9}$$

In Equation 9, $W_K, W_Q$, and $W_K$ are three linear transformation matrices to increase transformer's expressiveness, and $h$ is the dimension of the model.

### C. Proposed Model

In this paper, a hybrid bi-LSTM denoising autoencoder transformer model is presented for predicting hourly EV charging load. The structure of the hybrid model is illustrated in Figure 3.

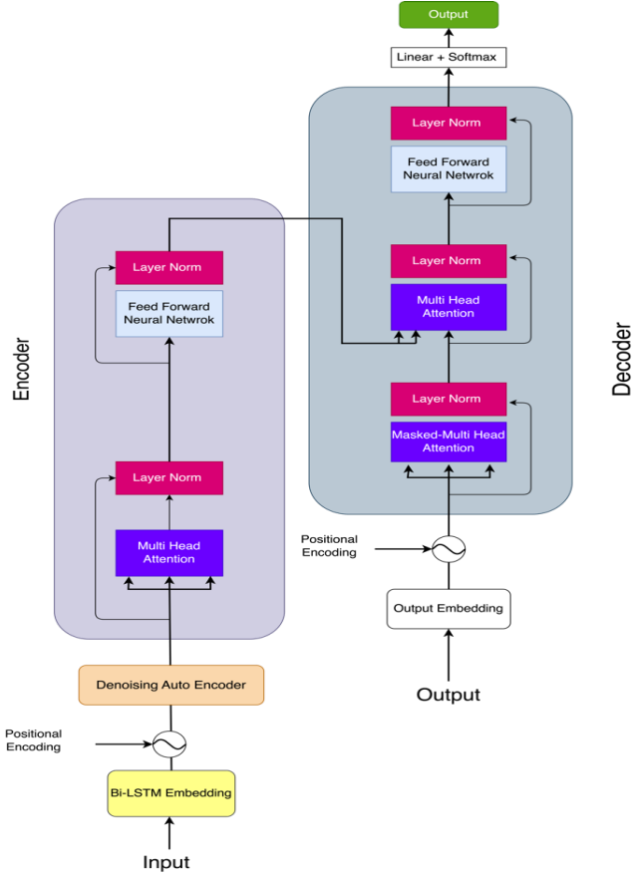

**Fig. 3.** Proposed Bi-LSTM embedding denoising auto encoder transformer.

The proposed model involves three main stages, each contributing to improving the accuracy of the predictions. The first stage of the proposed model involves using Bi-LSTM to embed the input data. By utilizing Bi-LSTM, the model can effectively extract relevant features from the input data, which can lead to improved predictions:

In the first stage of the modeling process, the input sequence is first introduced into the bi-directional long short-term memory (Bi-LSTM) network. This network comprises two LSTM models, with one processing the input sequence in a forward direction from left to right, and the other processing the sequence in reverse direction, from right to left. Next, the output from each of these LSTMs is a series of hidden states, which are then combined by concatenating them to form a final sequence of hidden states. Next, the final sequence of hidden states is passed through a fully connected layer, which produces a set of embeddings. These embeddings are then used as the input to the denoising autoencoder, which helps to refine the dataset by filtering out any extraneous noise or errors. This process helps to ensure that the data is accurate and reliable, which can lead to more precise and trustworthy predictions.

In the second stage, a denoising-auto encoder (DAE) is employed to filter out noise and errors present in the input data. The DAE helps to remove any unwanted or irrelevant information and hence helps to make the input data more reliable and accurate for the model's use. This filtering process results in a more precise and reliable dataset for the model to work with, which can lead to more accurate predictions.

The input to the DAE is the output of the Bi-LSTM embedding, which is a sequence of hidden states. The DAE applies a series of transformations to the input data to remove any unwanted or irrelevant information that might be present in the data. To achieve this, the DAE is trained to reconstruct the original input data after introducing random noise into it. By comparing the reconstructed data to the original data, the denoising autoencoder can identify which parts of the input data are important and which parts can be safely removed. The DAE then removes any extraneous information or errors that might be present in the input data.

The final stage of the proposed model involves using a transformer to process the filtered input data and make predictions about the hourly EV charging load. The transformer is a powerful neural network architecture that has been shown to be highly effective in many applications, such as natural language processing and image recognition [ref]. By incorporating the transformer into the model, the predictions can be made with even greater accuracy, allowing for better planning and management of EV charging infrastructure.

### D. Problem Statement

EV charging load prediction is modeled as a regression problem in modeling:

$$Y = f(X), \tag{10}$$

In this study, X denotes the historical data input, which corresponds to the EV charging load, while Y stands for the predicted value. To obtain this prediction, the study employs various regression functions denoted by F, such as LSTM, RNN, CNN, Transformer, and others.

Let $X = [x_1, x_2, ..., x_n]$ be the input time series data, where $x_i$ is the value of the time series at time i, and $T = [t_1, t, ..., t_n]$ be the corresponding timestamps. When creating time series embeddings using Bi-LSTM, the goal is to generate a vector representation for every element of the time series data that encompasses both its actual data value and corresponding timestamp. This is accomplished by combining the original time series data with a vector representation of its timestamp through concatenation. for each element of the time series data $x_i$, a corresponding time-value vector $t_{vi} = [t_i; \text{norm}(t_i)]$ is created where $t_i$ is the timestamp of $x_i$ and $\text{norm}(t_i)$ is the normalized value of $t_i$.

The input embedding E is created by concatenating the time-value vectors with the original time series data, where $E = [x_1; t_{v1}; x_2; t_{v2}; ...; x_n; t_{vn}]$.

The input data is initially fed into the Bi-LSTM embedding layer. the output of the Bi-LSTM layer is an embedded version of the input data, represented as embedded X, which is denoted as $X_{em}$.

$$X_{em} = Bi - LSTM(E),  \quad (11)$$

Bi-LSTM embedding structure is shown in figure 4:

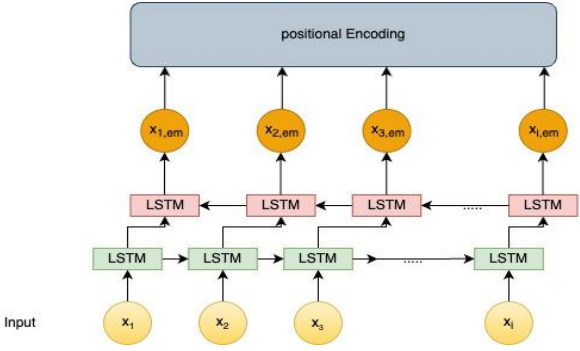

**Fig. 4.** Bi-LSTM Embedding architecture

The encoded time series generated by the Bi-LSTM embedding layer is shown as:

$$X_{em} = [x_{em}^1, x_{em}^2 \ldots, x_{em}^n],  \quad (12)$$

where $x_{em}^i$ is the encoded representation of the time series at time i. the next step is feeding the embedded input data corrupt the input, such as setting a fixed number of elements to zero or adding distributional noise, such as zero-mean Gaussian noise [26]. The corruption formula is written as:

$$\tilde{X}_{em} = f_c(X_{em}),  \quad (13)$$

where $\tilde{X}_{em}$ is the corrupted data, and $f_c(.)$ denotes the corrupt function. In the next step, the noisy data is fed to the encoder's neural network.

The encoder's purpose is to perform manifold learning, progressively decreasing the input data's dimensionality. Consequently, the latent values, which encapsulate the essential details of the original data and contain adequate information, are acquired.

For given data $(\tilde{X}_{em})$, the encoder function $f_\theta$ is expredded as:

$$f_\theta(\tilde{X}_{em}) = h = s\left(W_{\tilde{X}_{em}} + b\right),  \quad (14)$$

where W is the weight matrix, b is the bias vector, h is the latent value, and s ($\cdot$) is the activation function.

The decoder is the inverse of the encoder and is known as generative model learning which recovers the original data from the encoder's output. The decoder equation can be wrriten as:

$$g_{\hat\theta}(\text{h}) = \hat{X}_{em} = s\left(\acute{W}h + \acute{b}\right),  \quad (15)$$

where $\hat{X}_{em}$ represents the recovered data obtained from the input decoder or the encoder output of h, θ, and θ'.

The autoencoder learns the weight parameters θ, and θ' by minimizing a loss function $L(θ, θ')$, which quantifies the dissimilarity between the original input data and its corresponding output. Specifically, in this study, the mean-square error function is employed as the loss function to estimate the missing values in the time series data.

$$L(\theta, \theta') = \frac{1}{N}\sum_{k=1}^N \left\| X_{em}^k - \hat{X}_{em}^K \right\|^2 = \frac{1}{N}\sum_{k=1}^N \left\| X_{em}^k - g_{\hat\theta}(f_\theta(X_{em}^k)) \right\|^2,  \quad (16)$$

A DAE goes through three stages, including input corruption, encoding, and decoding. Figure 5 depicts the architecture of the DAE model.

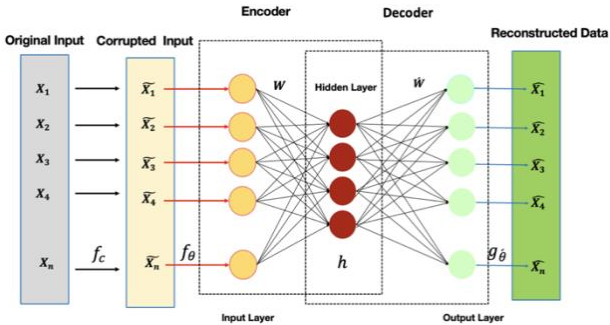

**Fig. 5.** Denoising auto encoder architecture

The input for the denoising autoencoder is created by adding noise to $X_{em}$. Various methods can be employed to

The $\hat{X}_{em}$ with lentgh *l*, and dimention *d*, is then transformed to sequences *query (Q), key (K),* and *value (V)*:

$$Q = \hat{X}_{em}W_Q,  \quad (17)$$

$$K = \hat{X}_{em}\,W_K,  \quad (18)$$

$$V = \hat{X}_{em}W_V  \quad (19)$$

where $W_K, W_Q,$ and $W_K$ are weighted matrices. Using $Q, K,$ and $V$, the self-attention score is calculated as:

$$A = Softmax(\frac{QK^T}{\sqrt{d_{qk}}})V,  \quad (20)$$

Multi-head attention then runs the attention mechanism several times in parallel for consistent performance improvements over conventional attention. The independent attention outputs are then concatenated and linearly transformed into the expected dimension.

### E. Training Objective

A MSE objective function was used during the training process of the models. A significant emphasis is placed on larger errors when evaluating the difference between the predicted and actual values. MSE can be expressed as follows:

$$MSE = \frac{\sum_{i=1}^{n}(\hat{y}_i - y_i)^2}{n},\qquad(21)$$

where $n$ is the number of the samples, $y_i$ and $\hat{y}_i$ are actual and predicted value at time step $i$, respectively.

### F. Hyperparameter Optimization

The optimization of hyperparameters plays a critical role in shaping the outcomes of deep learning algorithms by identifying the most effective parameter values. A hyperparameter is a set of predefined values used to enhance model performance, reduce overfitting, and streamline generalization prior to the learning process. However, exhaustively exploring the entire range of hyperparameter values can be time-consuming, particularly when dealing with larger parameter spaces. To address this, hyperparameters can be tuned using grid search and random search techniques. A manual search was conducted to optimize the hyperparameters. An overview of the hyperparameter range used in this study can be found in Table 1. During this process, multiple parameters were fine-tuned, revealing their significant impact on the model's accuracy.

Table1. Hyperparameter Value

| Hyper parameter | Value |
| --- | --- |
| Number of Layers | 1,3,6 |
| Number of Epoch | 10,50,100 |
| Number of Heads | 1,8 |
| Model Dimension | 32,64 |

## II. Guidelines For Manuscript Preparation

### G. Performance Measurement

Various performance measures are employed to assess the accuracy of models. When evaluating model performance in load forecasting, two commonly used error evaluation functions are the root mean squared error (RMSE) and the mean absolute error (MAE).

$$RMSE = \sqrt{\frac{\sum_{i=1}^{n}(\hat{y}_i - y_i)^2}{n}},\qquad(22)$$

The RMSE is utilized to penalize outliers and facilitate a clear interpretation of the forecasted output, as it is expressed in the same unit as the predicted feature. The equation for calculating RMSE is as follows:

We utilize MAE as an additional performance measure to improve and validate the reliability of the obtained values. The formula for MAE is as follows:

$$RMAE = \frac{\sum_{i=1}^{n}|\hat{y}_i - y_i|}{n},\qquad(23)$$

in the given formula, $n$ represents the number of samples, $y_i$ refers to the actual value, and $\hat{y}_i$ represents the forecasted value. A lower value of RMSE and MAE indicates superior prediction performance.

## III. EXPERIENCE AND ANALYSIS

### A. Data

The paper presents data on residential EV charging in Norwegian apartment buildings collected from December 2018 to January 2020. The data includes 6,878 individual charging sessions and was collected from a housing cooperative with 1,113 apartments and 2,321 residents in Norway, Europe [27]. Figure 6 presents the behavior of EV charging load at different time scales: hourly, weekly, and monthly. In Figure 6(a), the hourly EV charging load behavior is depicted throughout a 24-hour period, showing a significant increase from 4 pm to 8 pm. Figure 6 (b) showcases the EV charging load behavior specifically on weekdays, revealing higher levels during weekdays compared to weekend. In part (c) of Figure 6, the monthly EV charging load demonstrates a peak during the cold season. Lastly, Figure 7 illustrates the trend of EV charging load between the years 2019 and 2020.

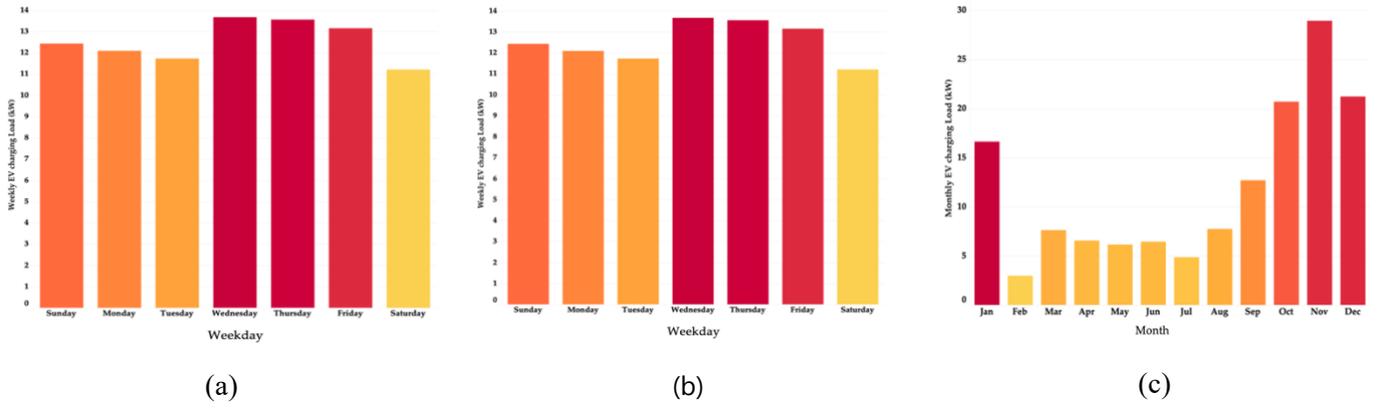

**Fig. 6.** EV charging load behavior at three different time scale: (a) hourly EV charging load, (b) weekly EV charging load, (c) Monthly EV charging load

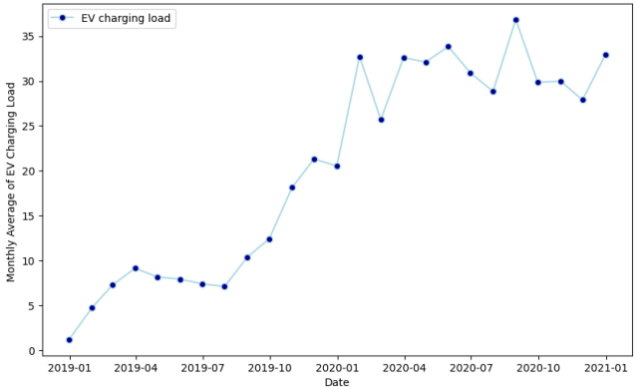

**Fig. 7.** EV charging load behavior at three different time scale

The pre-processed data is employed as input and split into three sets. The training set is comprised of 80% of the data, the validation set contains 10%, and the remaining 10% is assigned to the test set. The training set is shuffled before the training process begins. EV charging load is selected as the target parameters for forecasting sensory time series and are monitored throughout. The extensive multivariate time series data is utilized to build the model and evaluate the accuracy and dependability of the proposed method.

### B. Data Normalization

To ensure consistency and facilitate a smoother training process while avoiding large gradients, the data were normalized, resulting in each data point being on the same scale. A normalization procedure was performed on the datasets before they were fed into the models. A min-max normalization method was chosen for normalization. This type of normalization can be computed using the following formula:

$$x_{norm} = \frac{x_t - x_{min}}{x_{max} - x_{min}}, \tag{24}$$

where $x_{norm}$ is the normalized data, $x_t$ is the non-normalized data, and $x_{max}$ and $x_{min}$ are the maximum and minimum values of all the datasets, respectively.

IEEE will do the final formatting of your article. If your article is intended for a conference, please observe the conference page limits.

This is intended as an authoring template, not a final production template. It is not intended to match the final published format. Differences in final formatting are likely in the final IEEE files. Page count in the template is an estimate. Do not adjust line and character spacing to fit your paper to a specific length.

### C. Results

The proposed BDT model is evaluated against five benchmark models, namely Transformer, RNN, LSTM, CNN, and GRU. While the transformer and Bi-LSTM models have been previously described in Section 3, the other four benchmark models are briefly introduced as follows:

- RNN (Recurrent Neural Network): RNN is a type of neural network that operates on sequential data by maintaining a hidden state that captures information from previous time steps. It is widely used for modeling sequential dependencies in time series data [28].

- LSTM (Long Short-Term Memory): LSTM is a variant of RNN that is specifically designed to overcome the vanishing gradient problem and capture long-term dependencies. It utilizes a memory cell and different gating mechanisms to effectively handle sequential data [23].

- CNN (Convolutional Neural Network): CNN is a neural network architecture primarily used for image analysis, but it can also be adapted for time series analysis. It applies convolutional filters to extract local patterns and features from the input data [29].

- GRU (Gated Recurrent Unit): GRU is another variant of RNN that aims to address the limitations of traditional RNNs. It combines the gating mechanisms of LSTM but with a simplified architecture, making it computationally efficient while still capturing sequential dependencies [30].

These benchmark models serve as reference models for comparison with the proposed BDT model to assess its performance and effectiveness in time series forecasting tasks.

To obtain more comprehensive, and accurate findings, we executed the models multiple times to ensure that the results were not randomly generated. Additionally, we conducted a crucial examination by computing the standard deviation (STD) for both the RMSE and MAE for each time step. This significance test assists in determining whether observed disparities in evaluation outcomes are due to sampling error or mere chance.

The evaluation results of neural networks on the Hourly prediction of EV charging load are presented in Table 2, where the mean and standard error of the root mean squared error (RMSE) and mean absolute error (MAE) scores are reported for a total of five-time step runs. The prediction performance of each method is evaluated for forecasting horizons ranging from 24 to 120.

time steps, the BDT models achieved an RMSE of 0.145, 0.092, 0.100, 0.09, and 0.120, respectively, along with an MAE of 0.085, 0.066, 0.069, 0.06, and 0.089. These results indicate that the BDT model has significantly outperformed the other models, particularly in terms of MAE.

When compared to the second-best model, transformer, the proposed model has demonstrated superior performance in terms of MAE, with a significant reduction of 22.6%, 35.9%, 36%, and 56.1% for the 48 through 120-hour horizon.

This improved performance can be attributed to the proposed model's approach of combining Bi-LSTM architecture with denoising process.

It is important to note that in the case of the 24-hour forecasting horizon, the transformer, GRU, and LSTM models outperformed the proposed model. This could potentially be attributed to the fact that these models are more adept at capturing the short-term patterns and fluctuations in the EV charging load data, which are particularly important for forecasting over a 24-hour time frame.

On the other hand, the proposed Bi-LSTM denoising autoencoder transformer model is designed to incorporate more long-term dependencies in the data, which can better capture the complex and nonlinear

*Table2.* Performance comparison of deep learning models

| Time Horizon | Performance Measure | BDT | Transformer | RNN | LSTM | GRU | CNN |
|---|---|---|---|---|---|---|---|
| 24-h | RMSE | 0.145±0.02 | 0.083±0.04 | 0.552±0.01 | 0.13±0.001 | 0.102±0.009 | 0.512±0.002 |
| | MAE | 0.085±0.02 | 0.06±0.04 | 0.185±0.01 | 0.084±0.0008 | 0.098±0.002 | 0.39±0.00007 |
| 48-h | RMSE | **0.092±0.02** | 0.122±0.01 | 0.57±0.18 | 0.241±0.005 | 0.235±0.006 | 0.518±0.004 |
| | MAE | **0.066±0.01** | 0.103±0.02 | 0.211±0.21 | 0.159±0.003 | 0.154±0.004 | 0.387±0.002 |
| 72-h | RMSE | **0.100±0.01** | 0.122±0.01 | 0.543±0.002 | 0.314±0.009 | 0.256±0.004 | 0.548±0.003 |
| | MAE | **0.069±0.01** | 0.103±0.02 | 0.289±0.005 | 0.209±0.007 | 0.178±0.02 | 0.343±0.005 |
| 96-h | RMSE | **0.09±0.00** | 0.130±0.01 | 0.624±0.005 | 0.357±0.03 | 0.231±0.02 | 0.521±0.003 |
| | MAE | **0.06±0.00** | 0.110±0.01 | 0.33±0.009 | 0.238±0.001 | 0.188±0.001 | 0.384±0.00 |
| 120-h | RMSE | **0.120±0.001** | 0.165±0.01 | 0.609±0.001 | 0.367±0.005 | 0.23±0.003 | 0.609±0.02 |
| | MAE | **0.089±0.0005** | 0.139±0.01 | 0.342±0.0008 | 0.246±0.003 | 0.156±0.00 | 0.376±0.003 |
| | Total Win | 4 | 1 | 0 | 0 | 0 | 0 |

After analyzing the forecasting results across all horizons, it has been observed that the BDT method has outperformed all other models in four out of five time-horizons. The proposed model demonstrated superior performance in forecasting horizons ranging from 48 to 128 hours. Specifically, for the 24, 48, 72, 96, and 120-hour patterns present in the EV charging load data over longer forecasting horizons. This may explain why the proposed model consistently demonstrated superior performance in forecasting horizons ranging from 48 to 128 hours.

Overall, the evaluation results demonstrate the effectiveness of the proposed method in accurately predicting the EV charging load. These findings can serve as a valuable reference for researchers and practitioners working in the field of EV charging load prediction.

Figures 8 and 9 provide a visual representation of the RMSE and MAE values for the applied models, aiming to enhance understanding. These figures not only illustrate the magnitude of RMSE and MAE, but also demonstrate their variations over different time horizons. In Figure 8 and 9, it is evident that the proposed model exhibits a smooth line for both RMSE and MAE, indicating a balanced and stable performance across the entire time horizon. Conversely, the other applied models display more fluctuating lines, suggesting less stability in their performance as the number of time steps increases. Additionally, these models exhibit lower performance throughout the time steps.

charging load can be challenging due to the unpredictable nature of the data and the presence of noise. This study presents a novel approach to forecasting multistep short-term EV charging load, utilizing hourly EV charging history data. The proposed method is a hybrid system that combines Bi-LSTM embedding, denoising auto encoder, and transformer models. The system's innovations include the use of a non-linear Bi-LSTM model for embedding dataset which will lead to a more meaningful data representation, the implementation of a denoising auto encoder to effectively reduce noise in the original data and training the model using a vast amount of EV data. The system's performance is evaluated over three years of EV speed data.

The forecasting performance of different neural networks was evaluated using the RMSE and MAE

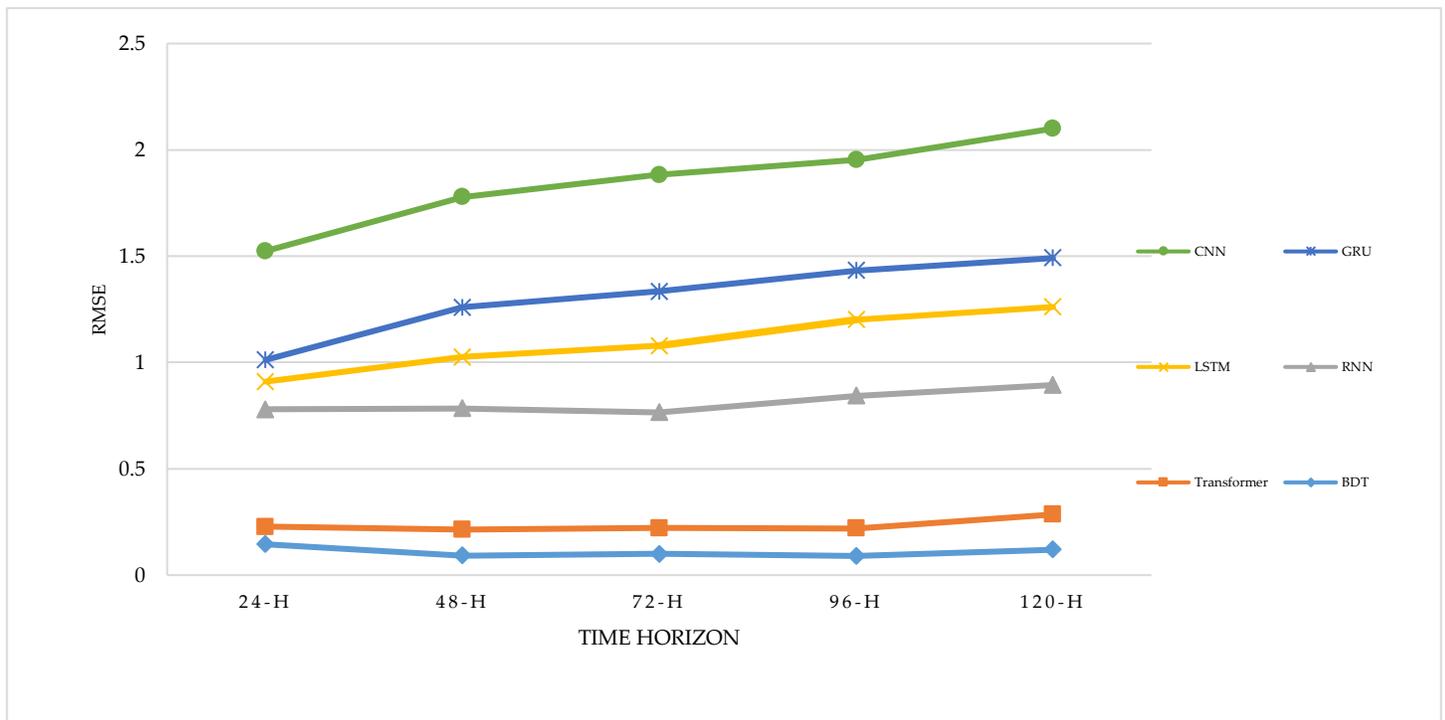

**Fig.7.** EV charging load behavior at three different time scale

## IV. SUMMARY AND CONCLUSION

Time series data predictions are essential in a wide variety of fields, including energy management. When it comes to electric vehicles (EVs), predicting the charging load accurately is crucial to balancing the grid and optimizing the use of energy. However, forecasting EV metrics for the prediction of EV charging load. This assessment was conducted across five forecasting horizons, spanning 24 to 120 hours. A bi-directional long short-term memory (Bi-LSTM) architecture was employed in combination with a denoising auto encoder process to develop our proposed model. The analysis showed that the proposed model outperformed all other models in 80% of the time (i.e., in four time horizons out of five), demonstrating superior performance in forecasting horizons ranging from 48 to 128 hours. The Bi-LSTM embedding denoising autoencoder transformer model is designed to incorporate more long-term dependencies in the data, better capturing the complex and nonlinear patterns present in the EV charging load data over longer forecasting horizons.

**Fig. 8.** EV charging load behavior at three different time scale

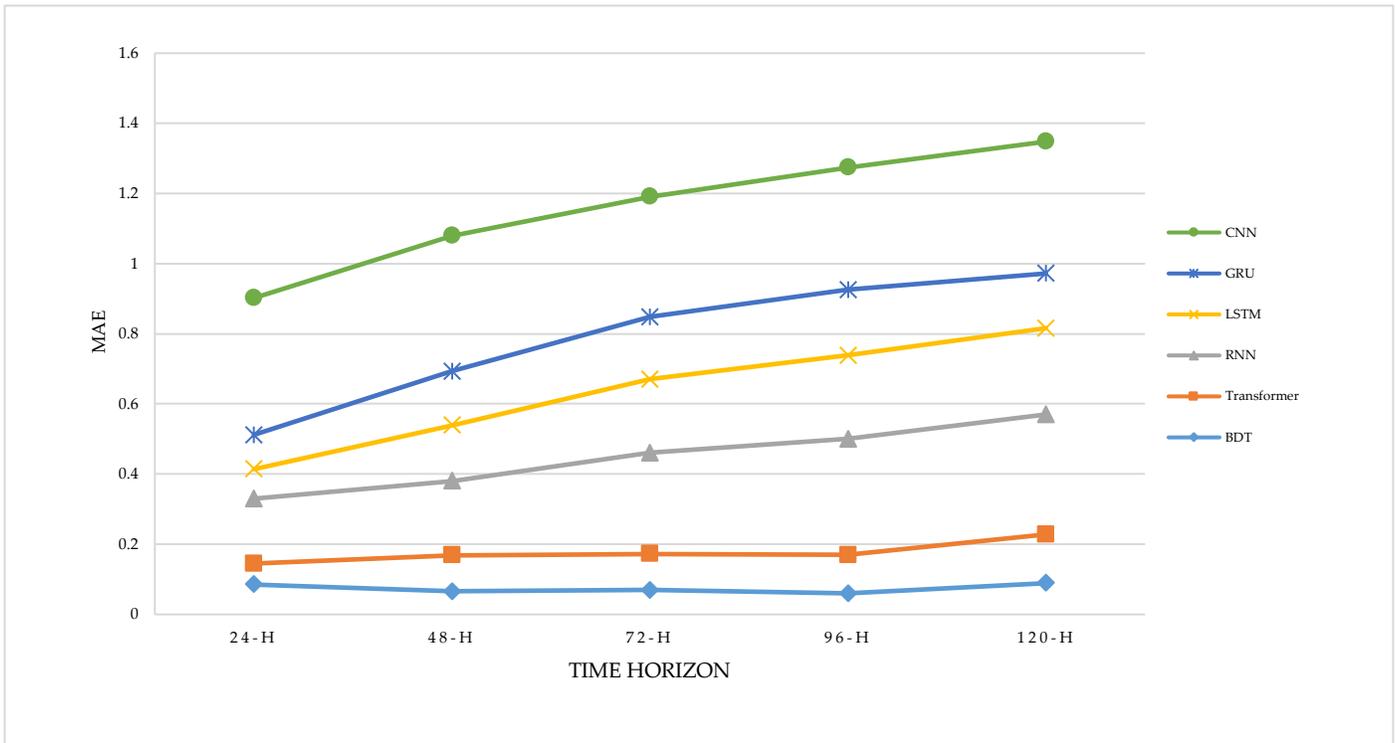

The accurate prediction of EV charging load is essential for effective energy management, and the proposed model has demonstrated high effectiveness in forecasting EV charging load. Our research outcomes can provide valuable insights for researchers and practitioners involved in the field of EV charging load prediction.

In future research, expanding the dataset in terms of quantity and input parameters, and examining the proposed model's effectiveness in different contexts such as energy sources can be explored.

## ACKNOWLEDGMENT

Authors want to acknowledge the contribution of Dr. Amit Kumar who has recently passed away.